\documentstyle[11pt,a4,epsf]{article}

\author{Hang Li \\
C\&C Research Laboratories, NEC Corporation \\
{\tt lihang@sbl.cl.nec.co.jp}}
\title{A Probabilistic Disambiguation Method Based on Psycholinguistic Principles}
\date{}

\begin{document}

\maketitle
\vspace{-1cm}

\begin{flushleft}
{\Large \bf Abstract}
\end{flushleft}
We address the problem of structural disambiguation in syntactic
parsing. In psycholinguistics, a number of principles of
disambiguation have been proposed, notably the Lexical Preference Rule
(LPR), the Right Association Principle (RAP), and the Attach Low and
Parallel Principle (ALPP). We argue that in order to improve
disambiguation results it is necessary to implement these principles
on the basis of a probabilistic methodology. We define a `three-word
probability' for implementing LPR, and a `length probability' for
implementing RAP and ALPP. Furthermore, we adopt the `back-off' method
to combine these two types of probabilities. Our experimental results
indicate our method to be effective, attaining an accuracy of
$89.2\%$.
\section{Introduction} 
Structural disambiguation is still a central problem in natural
language processing. To completely resolve ambiguities, we would need
to construct a human-like language {\em understanding} system
(c.f.\cite{Altmann88,Johnson-Laird83}). The construction of such a
system is extremely difficult, however, and we need to adopt a more
realistic approach. In psycholinguistics, a number of {\em principle}s
have been proposed which attempt to modelize the human disambiguation
process. The Lexical Preference Rule (LPR) \cite{Ford82}, the Right
Association Principle (RAP) \cite{Kimball73}, and the Attach Low and
Parallel Principle (ALPP, an extension of RAP) \cite{Hobbs90} have
been proposed, and it is thought that we might resolve ambiguities
quite satisfactorily if we could implement these principles
sufficiently \cite{Hobbs90,Whittemore90}. Methods of implementing
these principles have also been proposed (e.g.,
\cite{Shieber83,Wermter89,Wilks85}). An alternative approach is to
view language as a stochastic phenomenon, particularly from the
viewpoint of information theory and statistics.  If we could properly
define a probability model\footnote{A representation of a probability
  distribution is called a `probability model,' or simplely a
  `model.'} and calculate the likelihood value of each interpretation
using the model, we might also resolve ambiguities quite well. There
have been a number of methods proposed to perform structural
disambiguation using probability models, many of which have proved to
be quite effective
\cite{Alshawi95,Black92,Briscoe93,Chang92,Collins95,Fujisaki89,Hindle91,Hindle93,Jelinek90,Magerman91,Magerman95,Ratnaparkhi94,Resnik93a}\\ \cite{Su88}.

Although each of the disambiguation methods proposed to date has its
merits, none resolves the disambiguation problem completely
satisfactorily. We feel that it is necessary to devise a new method
that unifies the above two approaches, i.e., to implement
psycholinguistic principles of disambiguation on the basis of a
probabilistic methodology. Most psycholinguistic principles have been
developed on the basis of vast data of actual observations, and thus a
method based on them is expected to achieve good disambiguation
results.  Probabilistic methods of implementing these principles have
the merit of being able to handle noisy data, as well as being able to
employ a principled methodology for acquiring the knowledge necessary
for disambiguation.

LPR, RAP and ALPP are known to be effective for disambiguation, and
these are the ones whose implementation we consider in the present
paper. Thus our problem involves the following three subproblems: (a)
resolving structural ambiguities based on LPR in terms of
probabilistic representations, (b) resolving structural ambiguities
based on RAP and ALPP in terms of probabilistic representations, and
(c) combining the two. For subproblem (a), we have devised a new
method, based on LPR, which has some good properties not shared by the
methods proposed so far
\cite{Alshawi95,Chang92,Collins95,Hindle91,Ratnaparkhi94,Resnik93a}.
In \cite{Li95}, we have described this method in detail. In the
present paper, we mainly describe our solutions to subproblems (b) and
(c). For subproblem (b), we point out that the notion of the `length'
of a syntactic category \footnote{The length of a syntactic category
  in simply defined as the number of words contained in that
  category.} is important, and propose to use a `length probability'
to perform structural disambiguation. For subproblem (c), we propose
to adopt the `back-off' method, i.e., to make use first of a lexical
likelihood based on LPR, and then a syntactic likelihood based on RAP
and ALPP. Experiments conducted to test the effectiveness of our
method demonstrate an encouraging accuracy of $89.2\%$.
\section{Psycholinguistic Principles of Disambiguation}\label{sec:psycho}
In this section, we introduce the psycholinguistic principles of
disambiguation. Kimball has proposed the Right Association Principle
(RAP) \cite{Kimball73}, which states that (in English) a phrase on the
right should be attached to the nearest phrase on the left if
possible. Hobbs \& Bear have generalized RAP to the Attach Low and
Parallel Principle (ALPP) \cite{Hobbs90}. ALPP states that a phrase on
the right should be attached to the nearest phrase on the left if
possible, and that phrases should be attached to a phrase in parallel
if possible. (When we refer to ALPP, we ordinarily mean just the part
concerning attachments in parallel. ) Ford~{\it et~al.} have proposed
the Lexical Preference Rule (LPR) which states that an interpretation
is to be preferred whose case frame assumes more semantically
consistent values \cite{Ford82}. Classically, lexical preference is
realized by checking consistencies between `semantic features' of
slots and those of slot values, namely the `selectional restrictions'
\cite{Katz63}. The realization of lexical preference in terms of
selectional restrictions has some disadvantages, however.
Interpretations obtained in an analysis cannot, for example, be ranked
in their preferential order. Thus one cannot adopt a strategy of
always retaining the $N$ most plausible partial interpretations in an
analysis, which is the most widely accepted practice at present. In
fact it is more appropriate to treat the lexical preference as a kind
of score representing the association between slots and their values.
In the present paper, we refer to this kind of score as `lexical
preference.'  For the same reason, we also treat `syntactic
preference' as a kind of score.

LPR is a lexical semantic principle, while RAP and ALPP are syntactic
ones, and in psycholinguistics it is commonly claimed that LPR
overrides RAP and ALPP \cite{Hobbs90}. Let us consider some examples
of LPR and RAP in this regard. For the sentence
\begin{equation}\label{eq:eat}
 \mbox{I ate ice cream with a spoon,}
\end{equation}
there are two interpretations; one is `I ate ice cream using a spoon'
and the other `I ate ice cream and a spoon.' In this sentence, a human
speaker would certainly assume the former interpretation over the
latter.  From the psycholinguistic perspective, this can be explained
in the following way: the former interpretation has a stronger lexical
preference than the latter, and thus is to be preferred according to
LPR. Moreover, since LPR overrides RAP, the preference is solely
determined by LPR. For the sentence
\begin{equation}\label{eq:phone}
\mbox{John phoned a man in Chicago,}
\end{equation}
there are two interpretations; one is `John phoned a man who is in
Chicago' and the other `John, while in Chicago, phoned a man.' In this
sentence, a human speaker would probably assume the former
interpretation over the latter. The two interpretations have an equal
lexical preference value, and thus the preference of the two cannot be
determined by LPR. After LPR fails to work, the former interpretation
is to be preferred according to RAP, because `a man' is closer to `in
Chicago' than `phone' in the sentence.

LPR implies that (in natural language) one should communicate as
relevantly as possible, while RAP and ALPP implies that one should
communicate as efficiently as possible. Although the phenomena
governed by these principles vary from language to language, the
principles themselves, we think, are {\em language independent}, and
thus can be regarded as fundamental principles of human communication.
According to Whittemore~{\it et~al.} and Hobbs \& Bear, nearly all of
the ambiguities can be resolved by first applying LPR and then RAP and
ALPP \cite{Hobbs90,Whittemore90}. These observations motivate us
strongly to implement these principles for disambiguation purposes.

While there are also other principles proposed in the literature,
including the Minimal Attachment Principle \cite{Frazier79}, they are
generally either not highly functional or covered by the above three
principles in any case \cite{Hobbs90,Whittemore90}.

The necessity of developing a disambiguation method with learning
ability has recently come to be widely recognized. The realization of
such a method would make it possible to (a) save the cost of defining
knowledge by hand (b) do away with the subjectivity inherent in human
definition (c) make it easier to adapt a natural language analysis
system to a new domain. We think that a probabilistic approach is
especially attractive because it is able to employ a principled
methodology for acquiring the knowledge necessary for disambiguation.
In our research, we implement LPR, RAP and ALPP by means of a
probabilistic methodology.
\section{LPR and Lexical Likelihood}
In this section, we briefly describe our LPR-based probabilistic
disambiguation method.
\subsection{The three-word probability}
We refer to a syntactic tree and its corresponding case frame, as
obtained in an analysis, `an interpretation.'\footnote{We do not take
 into account ambiguities caused by word senses.} After analyzing the
sentence in (\ref{eq:eat}), for example, we obtain the case frames of
the interpretations:
\begin{equation}\label{eq:eatframe1}
 \mbox{eat:[arg1 I, arg2 ice\_cream, with spoon]},
\end{equation}
and
\begin{equation}\label{eq:eatframe2}
\mbox{eat:[arg1 I, arg2 ice\_cream: [with spoon]].}
\end{equation}
The value assumed by a case slot of a case frame of a verb can be
viewed as being generated according a conditional probability
distribution:
\begin{equation}
P(n | v, s),
\end{equation}
where random variable $v$ takes on a value of a set of verbs, $n$ a
value of a set of nouns, and $s$ a value of a set of slot names.
Similarly, the value assumed by a case slot of a case frame of a noun
can be viewed as being generated by a conditional probability
distribution: $P(n|n,s)$. We call this kind of conditional probability
the `three-word probability.' Moreover, we assume that the three-word
probabilities in the case frame of an interpretation are mutually
independent, and define the geometric mean of the three-word
probabilities as the `lexical likelihood' of the interpretation:
\begin{equation}
P_{lex}(I) = (\prod_{i=1}^{m} P_i)^{1/m}, 
\end{equation}
where $P_i$ is the $i$th three-word probability in the case frame of
interpretation $I$, and $m$ the number of three-word probabilities in
it. The lexical likelihood values of the two interpretations in
(\ref{eq:eatframe1}) and (\ref{eq:eatframe2}) are thus calculated as
\begin{equation}
\begin{array}{ll}
 P_{lex}( I_1 ) = & (P({\rm I}|{\rm eat},{\rm arg1})\times P({\rm
 ice\_cream}|{\rm eat},{\rm arg2})\times P({\rm spoon}|{\rm eat},{\rm
 with}))^{1/3},
\end{array}
\end{equation}
and
\begin{equation}
\begin{array}{ll}
 P_{lex}( I_2 ) = & (P({\rm I}|{\rm eat},{\rm arg1})\times P({\rm
 ice\_cream}|{\rm eat},{\rm arg2})\times P({\rm
 spoon}|{\rm ice\_cream},{\rm with}))^{1/3}.
\end{array}
\end{equation}
In disambiguation, we simply rank the interpretations according to
their lexical likelihood values. If a verb (or a noun) has a strong
tendency to require a certain noun as the value of its case frame
slot, the estimated three-word probability for such a co-currence will
be very high. To prefer an interpretation with a higher lexical
likelihood value, then, is to prefer it based on its lexical
preference.  Specifically, in order to perform pp-attachment
disambiguation in analysis of sentences like (\ref{eq:eat}), we need
only calculate and compare the values of $P({\rm spoon}|{\rm eat},{\rm
  with})$ and $P({\rm spoon}|{\rm ice\_cream},{\rm with})$. In
sentences like
\begin{equation}
 \mbox{A number of companies sell and buy by computer,}
\end{equation}
the number of three-word probabilities in each of its respective
interpretations will be different. If we were to define a lexical
likelihood as the product of the three-word probabilities in the case
frame of an interpretation, an interpretation with fewer case slots
would be preferred. We use the definition of lexical likelihood
described above to avoid this problem.\footnote{An alternative for
  resolving this kind of ambiguity (coordinate structure ambiguity) is
  to employ a method which examines the similarity that exists between
  conjuncts (c.f.\cite{Kurohashi94,Resnik93a}).}
\subsection{The data sparseness problem}
Hindle \& Rooth have previously proposed resolving pp-attachment
ambiguities with `two-word probabilities' \cite{Hindle91}, e.g.,
$P({\rm with}|{\rm ice\_cream}),P({\rm with}|{\rm eat})$, but these are
not accurate enough to represent lexical preference. For example, in
the sentences,
\begin{equation}
\begin{array}{l}
 \mbox{Britain reopened the embassy in December,} \\ \mbox{Britain
 reopened the embassy in Teheran,}
\end{array}
\end{equation}
the pp-attachment sites of the two prepositional phrases are
different. The attachment sites would be determined to be the same,
however, if we were to use two-word probabilities
(c.f.\cite{Resnik93a}), and thus the ambiguity of only one of the
sentences can be resolved. It is very likely, however, that this kind
of ambiguity could be resolved satisfactorily by using the three-word
probabilities.

The number of parameters that need to be estimated increases
drastically when we use three-word probabilities, and the data
available for estimation of the probability parameters usually are not
sufficient in practice. If we employ the Maximum Likelihood Estimator,
we may find most of the parameters are estimated to be $0$: a problem
often referred to, in statistical natural language processing, as the
`data sparseness problem.' (The motivation for using the two-word
probabilities in \cite{Hindle91} appears to be a desire to avoid the
data sparseness problem.) One may expect this problem to be less
severe in the future, when more data are available. However, as data
size increases, new words may appear, and the number of parameters
that need to be estimated may increase as well. Thus, the data
sparseness problem is unlikely to be resolved. A number of methods
have been proposed, however, to cope with the data sparseness problem.
Chang~{\it et~al.}, for instance, have proposed replacing words with
word classes and using class-based co-occurrence probabilities
\cite{Chang92}. However, forcibly replacing words with certain word
classes is too loose an approximation, which, in practice, could
seriously degrade disambiguation results. Resnik has defined a
probabilistic measure called `selectional association' in terms of the
word classes existing in a given thesaurus. While Resnik's method is
based on an interesting intuition, the justification of this method
from the viewpoint of statistics is still not clear. We have devised a
method of estimating the three-word probabilities in an efficient and
theoretically sound way \cite{Li95}. Our method selects optimal word
classes according to the distribution of given data, and smoothes the
three-word probabilities using the selected classes. Experimental
results indicate that our method improves upon or is at least as
effective as existing methods. Using our method of estimating
(smoothing) probabilities, we can cope with the data sparseness
problem. However, for the same reason as described above, the data
sparseness problem cannot be resolved completely. We propose combining
the use of three-word probabilities and that of two-word
probabilities. Specifically, we first use the lexical likelihood value
calculated as the geometric mean of the three-word probabilities of an
interpretation; and when the lexical likelihood values of obtained
interpretations are equal, including the case in which all of them are
$0$, we use the lexical likelihood value calculated as the geometric
mean of the two-word probabilities of an interpretation.
\section{RAP,ALPP, and Syntactic Likelihood}\label{sec:rap}
In this section, we describe our probabilistic disambiguation method
based on RAP and ALPP.
\subsection{The deterministic approach}
Shieber has previously proposed incorporating RAP into the mechanism
of a shift-reduce parser \cite{Shieber83}. When RAP is implemented,
the parser prefers shift to reduce whenever a `shift-reduce conflict'
occurs. The advantage of this deterministic approach is its simple
mechanism, while the disadvantage is that although it can output the
most preferred interpretation, it cannot rank interpretations in their
preferential order. In order to be able to rank interpretations in
this way, it is necessary to construct a parser which operates
stochastically, not deterministically.
\subsection{Formalizing a syntactic preference}
In this subsection, we formalize a syntactic preference based on RAP
and ALPP.
\begin{figure}[htb]
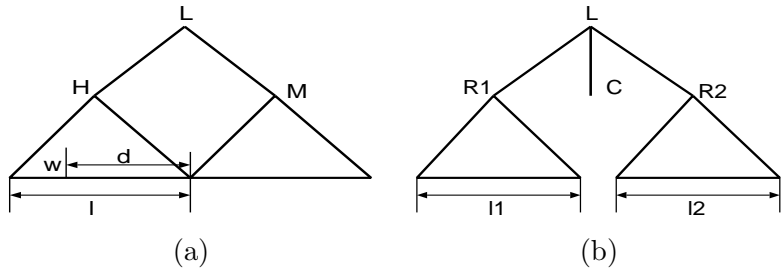

\begin{center}
\begin{tabular}{cc}
{\epsfxsize5cm\epsfysize3cm\epsfbox{raplen.eps}} &
{\epsfxsize5cm\epsfysize3cm\epsfbox{alpplen.eps}} \\
(a) & (b) \\
\end{tabular}
\end{center}
\caption{RAP, ALPP and length}
\label{fig:length}
\end{figure}
While we borrow from the terminology of HPSG \cite{Pollard87} in our
reference to `head' categories, we also use the term `modifier'
categories to refer to categories which HPSG would classify as being
either `complements' or `adjuncts.' We refer to that word which
exhibits the subcategory feature of a category to be that category's
`head word.'

Let us consider a simple case in which we are dealing with a modifier
category $M$, a head category $H$, and the head word of $H$, $w$. We
first apply CFG rule $L \rightarrow H,M$ to $H$ and $M$, yielding
category $L$ (see Figure~\ref{fig:length}(a)). We refer to the number
of words in a given sequence as `distance.' As may be seen in
Figure~\ref{fig:length}(a), the distance between $M$ and $w$ is $d$.
RAP prefers an interpretation with a smaller $d$. Thus, syntactic
preference can be represented by a monotonically decreasing function
of $d$. Since in English the head word $w$ of category $H$ tends to
locate near its left corner, we can approximate $d$ as $l$, the number
of words contained in $H$. In this paper, we call the number of words
contained in a category the `length' of that category. In addition,
syntactic preference also depends on type of head category and
modifier category. Assume that $l$ is known to be $5$; if $H$ is a
verb phrase and $M$ is a prepositional phrase, the preference value is
likely to be high, but if $H$ is a noun phrase and $M$ is a
prepositional phrase, it is likely to be low. Since category type can
be specified within a CFG rule, syntactic preference can be defined as
a function of a CFG rule. Syntactic preference based on RAP can be
formalized, then, as a function of CFG rule $L \rightarrow H,M$ and
length $l$, namely,
\begin{equation}\label{eq:scorerap}
S(l,(L \rightarrow H,M)).
\end{equation}

Suppose that categories $R_1$ and $R_2$ form a coordinate structure,
and $l_1$ and $l_2$ are the lengths of $R_1$ and $R_2$, respectively.
ALPP prefers categories forming a coordinate structure to be of equal
length (see Figure~\ref{fig:length}(b)). Preference value will be high
when $l_1$ equals $l_2$, and syntactic preference based on
ALPP\footnote{This kind of syntactic preference requires that the CFG
 rules for coordinate structures have the form $L \rightarrow
 R_1,C,R_2,C,\ldots,C,R_k$.} can be defined as
\begin{equation}\label{eq:scorealpp}
S(l_1,l_2,(L \rightarrow R_1,C,R_2)).
\end{equation}

Further, suppose that categories $R_1, R_2, \ldots, R_k$ are combined
into category $A$, and $l_1,l_2,\ldots,l_k$ are the lengths of $R_1,
R_2, \ldots, R_k$, respectively. Syntactic preference of the
attachment can then be defined as
\begin{equation}\label{eq:scoregen}
 S(l_1,l_2,\ldots,l_k,(L \rightarrow R_1,R_2,\ldots,R_k)).
\end{equation}
Note that (\ref{eq:scoregen}) contains (\ref{eq:scorerap}) and
(\ref{eq:scorealpp}). Furthermore, we assume that the attachments in
the syntactic tree of an interpretation are mutually independent, and
we define the product (or the sum, depending on the preference
function) of the syntactic preference values of the attachments in the
syntactic tree of the interpretation as the syntactic preference of
the interpretation:
\begin{equation}
 S_{syn}(I) = \prod_{i=1}^{m} S_i,
\end{equation}
where $S_i$ denotes the syntactic preference value of the $i$th
attachment in the syntactic tree of interpretation $I$, and $m$ the
number of attachments in it.
\subsection{The length probability}
We now consider how to specify the syntactic preference function in
(\ref{eq:scoregen}). As there are any number of ways to formulate the
function (note the fact that syntactic preference is also a function
of a CFG rule.), it is nearly impossible to find the most suitable
formula experimentally. To cope with this problem, we used machine
learning techniques (recall the merits of using machine learning
techniques in disambiguation, as described in
Section~\ref{sec:psycho}). Specifically, we have defined a probability
model to calculate syntactic preference. Suppose that attachments
represented by CFG rules and lengths are extracted from the {\em
 correct} syntactic trees in training data, and the frequency of each
kind of attachment is obtained as
\begin{equation}\label{eq:freqattach}
f(l_1,l_2,\ldots,l_k,(L \rightarrow R_1,R_2,\ldots,R_k)),
\end{equation}
where $L \rightarrow R_1,R_2,\ldots,R_k$ denotes a CFG rule, and
$l_1,l_2,\ldots,l_k$ denote the lengths of $R_1,R_2,\ldots,R_k$,
respectively. RAP prefers an interpretation attached to a nearer
phrase, while ALPP prefers interpretations with attachments that are
low and in parallel. Many such attachments may be observed in the
training data, and we can formulate the frequencies of attachments
(\ref{eq:freqattach}) as a syntactic preference. Considering the fact
that individual rules will be applied with different frequency, it is
desirable to modify the syntactic preference to
\begin{equation}
 \frac{f(l_1,l_2,\ldots,l_k,(L \rightarrow R_1,R_2,\ldots,R_k))}
 {f((L \rightarrow R_1,R_2,\ldots,R_k))},
\end{equation}
where $f((L \rightarrow R_1,R_2,\ldots,R_k))$ denotes the frequence of
application of CFG rule $L \rightarrow R_1,R_2,\ldots,R_k$. This is
precisely the `length probability' we propose in this paper.

Let us now define the length probability more formally. Suppose that
an attachment is obtained after the application of CFG rule $L
\rightarrow R_1,R_2,\ldots,R_k$, the lengths of $R_1,R_2,\ldots,R_k$
are $l_1,l_2,\ldots,l_k$, respectively. The attachment can be viewed
as being generated by the following conditional distribution:
\begin{equation}\label{eq:disprob}
P(l_1,l_2,\ldots,l_k|(L \rightarrow R_1,R_2,\ldots,R_k)).
\end{equation}
We call this kind of conditional probability the `length probability.'
\footnote{The number of parameters in a length probability model
 depends on $k$ - the number of categories on the right-hand side of
 a CFG rule, and $N$ - the maximum value of lengths of a category on
 the left-hand side of the rule:
 $\sum_{i=k-1}^{N-1} \left(
 \begin{array}{c} i \\ k-1
\end{array}\right) - 1 = \left(\begin{array}{c} N \\ k
\end{array}\right) - 1.$
As $k$ is very small (in our case $k\le 3$), the number of parameters
in a length probability model is of $N$'s polynomial order.}
Furthermore, the syntactic likelihood of an interpretation is defined
as the geometric mean of the length probabilities of the attachments
in the syntactic tree of the interpretation, assuming that the
attachments are mutually independent:
\begin{equation}
P_{syn}(I) = (\prod_{i=1}^{m} P_i)^{\frac{1}{m}},
\end{equation}
where $P_i$ is the $i$th length probability in the syntactic tree of
interpretation $I$, and $m$ the number of length probabilities in it.
We define syntactic likelihood as the geometric mean of the length
probabilities, rather than as the product of the length probabilities,
in order to factor out the effect of the different number of
attachments in the syntactic trees of individual interpretations. When
training the length probabilities, the parameters in
(\ref{eq:disprob}) may be estimated using the frequences in
($\ref{eq:freqattach}$).

\begin{figure}[htb]
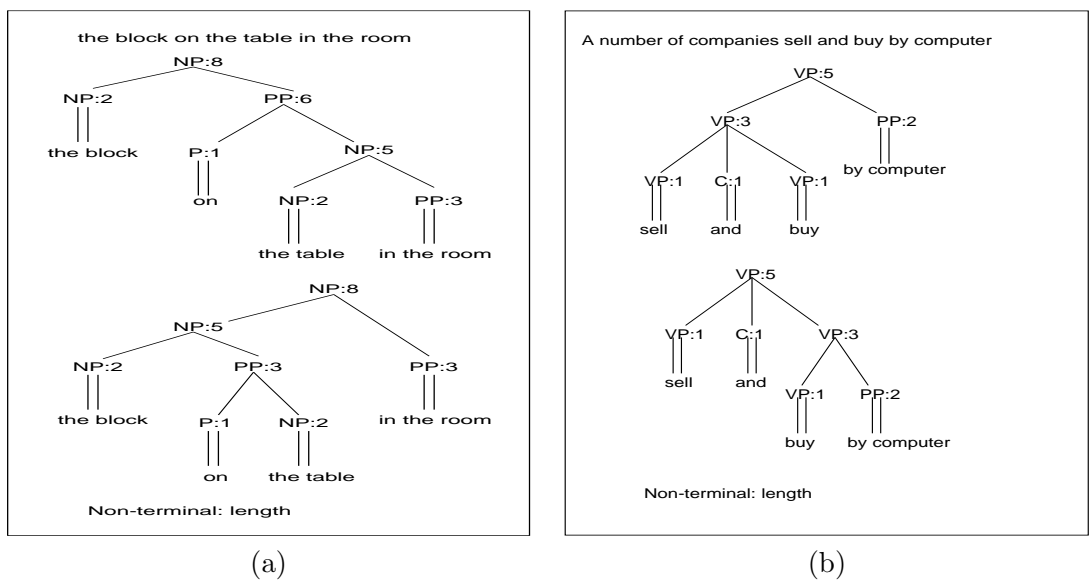

\begin{center}
\begin{tabular}{cc}
{\epsfxsize7cm\epsfysize7cm\epsfbox{rapex.eps}} &
{\epsfxsize7cm\epsfysize7cm\epsfbox{alppex.eps}} \\
(a) & (b) \\
\end{tabular}
\caption{Examples of syntactic parsing}
\label{fig:example}
\end{center}
\end{figure}

Next, let us consider a simple example illustrating how the operation
of this model indicates the functioning of RAP. For the phrase shown
in Figure~\ref{fig:example}(a), there are two interpretations; RAP
would necessarily prefer the former. The difference between the
syntactic likelihood values of the two interpretations is solely
determined by
\begin{equation}\label{eq:blockprob1}
 P(1,5|(PP \rightarrow P,NP))\times P(2,6|(NP \rightarrow NP,PP)),
\end{equation}
and
\begin{equation}\label{eq:blockprob2}
 P(1,2|(PP \rightarrow P,NP))\times P(5,3|(NP \rightarrow NP,PP)).
\end{equation}
First, let us compare the left-hand length probabilities of
(\ref{eq:blockprob1}) and (\ref{eq:blockprob2}). Both represent an
attachment of $NP$ to $P$, and the length of $P$ is $1$ in both terms.
Thus the two estimated probabilities may not differ so greatly. Next,
compare the right-hand length probabilities in (\ref{eq:blockprob1})
and (\ref{eq:blockprob2}). While both represent an attachment of $PP$
to $NP$, the length of $NP$ of the former is $2$ and that of the
latter is $5$. Thus the second length probability in
(\ref{eq:blockprob1}) is likely to be higher than that in
(\ref{eq:blockprob2}), as in training data there are more phrases
attached to nearby phrases than are attached to distant ones.
Therefore, when we use only the syntactic likelihood to perform
disambiguation, we can expect the former interpretation in
Figure~\ref{fig:example}(a) to be preferred, i.e., we have an
indication of the functioning of RAP.

Let us consider another example illustrating how the operation of the
length probability model indicates the functioning of ALPP. For the
sentence shown in Figure~\ref{fig:example}(b), there are two
interpretations; ALPP would necessarily prefer the former. The
difference between the syntactic likelihood values of the two
interpretations is solely determined by
\begin{equation}\label{eq:compprob1}
 P(3,2|(VP \rightarrow VP,PP))\times P(1,1,1|(VP \rightarrow VP,C,VP)),
\end{equation}
and
\begin{equation}\label{eq:compprob2}
 P(1,2|(VP \rightarrow VP,PP))\times P(1,1,3|(VP \rightarrow VP,C,VP)).
\end{equation}
First, let us compare the left-hand length probabilities in
(\ref{eq:compprob1}) and (\ref{eq:compprob2}). Both represent an
attachment of $PP$ to $VP$, but the length of $VP$ of the former is
$3$ and that of the latter is $1$. The left-hand probability in
(\ref{eq:compprob1}) is likely to be lower than that in
(\ref{eq:compprob2}). Next, compare the right-hand length
probabilities in (\ref{eq:compprob1}) and (\ref{eq:compprob2}). Both
represent a coordinate structure consisting of $VP$s. The lengths of
$VP$s in the latter are equal, while the lengths of $VP$s in the
former are not. Thus the right-hand probability in
(\ref{eq:compprob1}) is likely to be higher than that in
(\ref{eq:compprob2}). Moreover, the difference between the right-hand
probabilities is likely to be higher than that between the left-hand
probabilities, and thus the syntactic likelihood value of the former
interpretation will be higher than that of the latter. Therefore, when
we use only the syntactic likelihood to perform disambiguation, we can
expect the former interpretation in Figure~\ref{fig:example}(b) to be
preferred.
\subsection{The syntactic parsing approach}
Another approach to disambiguation is to define a probability model on
the basis of syntactic parsing. One method of this type employs the
well-known PCFG (Probabilistic Context Free Grammar) model
\cite{Fujisaki89,Jelinek90,Lari90}. In PCFG, a CFG rule having the
form of $\alpha \rightarrow \beta$ is associated with a conditional
probability $P(\beta |\alpha )$, and the likelihood of a syntactic
tree is defined as the product of the conditional probabilities of the
rules which are applied in the derivation of that tree. Other methods
have also been proposed. Magerman \& Marcus, for instance, have
proposed making use of a conditional probability model specifying a
conditional probability of a CFG rule, given the part-of-speech
trigram it dominates and its parent rule \cite{Magerman91}. Black~{\it
  et~al.} have defined a richer model to utilize all the information
in the top-down derivation of a non-terminal \cite{Black92}. Briscoe
\& Carroll have proposed using a probabilistic model specific to LR
parsing \cite{Briscoe93}.

The advantage of the syntactic parsing approach is that it may embody
heuristics (principles) effective in disambiguation, which would not
have been thought of by humans, but it also risks not embodying
heuristics (principles) already known to be effective in
disambiguation. For example, the two interpretations of the noun
phrase shown in Figure~\ref{fig:example}(a) have an equal likelihood
value, if we employ PCFG, although the former would be preferred
according to RAP.
\section{The Back-Off Method}
Having defined a lexical likelihood based on LPR and a syntactic
likelihood based on RAP and ALPP, we may next consider how to combine
the two kinds of likelihood in disambiguation. One choice is to
calculate total preference as a weighted average of likelihood values,
as proposed in \cite{Alshawi95}. However since LPR overrides RAP and
ALPP, a simpler approach is to adopt the back-off method, i.e., to
rank interpretations $I_1$ and $I_2$ as follows:
\begin{equation}\label{eq:pref}
\begin{array}{lllll}
1. & \mbox{if} & P_{lex}(I_1) - P_{lex}(I_2) > \eta & \mbox{then} & I_1 > I_2 \\
2. & \mbox{else if} & P_{lex}(I_2) - P_{lex}(I_1) > \eta & \mbox{then} & I_2 > I_1 \\
3. & \mbox{else if} & P_{syn}(I_1) - P_{syn}(I_2) > \tau & \mbox{then} & I_1 > I_2 \\
4. & \mbox{else if} & P_{syn}(I_2) - P_{syn}(I_1) > \tau & \mbox{then} & I_2 > I_1 \\
\end{array}
\end{equation}
where $I_1$ and $I_2$ denote any two interpretations, $P_{lex}()$
denotes the lexical likelihood of an interpretation, and $P_{syn}()$
the syntactic likelihood of an interpretation. $\eta \ge 0$ and $\tau
\ge 0$ are thresholds (in the experiment described later, both are set
to $0$). Note that in lines 3 and 4,
$|P_{lex}(I_1)-P_{lex}(I_2)|\le\eta$ holds. Further note that the
preferential order cannot be determined (or can only be determined at
random) when $|P_{lex}(I_1)-P_{lex}(I_2)|\le\eta$ and
$|P_{syn}(I_1)-P_{syn}(I_2)|\le\tau$.
\section{Experimental Results}
We have conducted experiments to test the effectiveness of our
proposed method. This section describes the results. In the
experiments, we considered only resolving pp-attachment ambiguities and
coordinate structure ambiguities. These two kinds of ambiguities are
typical, and other ambiguities can be resolved in the same way
\cite{Hobbs90}.

We first defined 12 CFG rules as our grammar to be used by a parser
which calculates a preference for each partial interpretation, and
always retains the $N$ most preferable partial
interpretations\footnote{It is necessary to do so, as the number of
  ambiguities will increase drastically when the length of an input
  sentence increases \cite{Church82}. }. We have not yet actually
constructed such a parser, however, and use a parser called `SAX,'
previously developed by Matsumoto \& Sugimura \cite{Matsumoto86e},
which calculates a preference for each interpretation after it obtains
all the interpretations.

\begin{figure}[htb]
\begin{center}
\begin{tabular}{cc}
{\epsfxsize6cm\epsfysize4cm\epsfbox{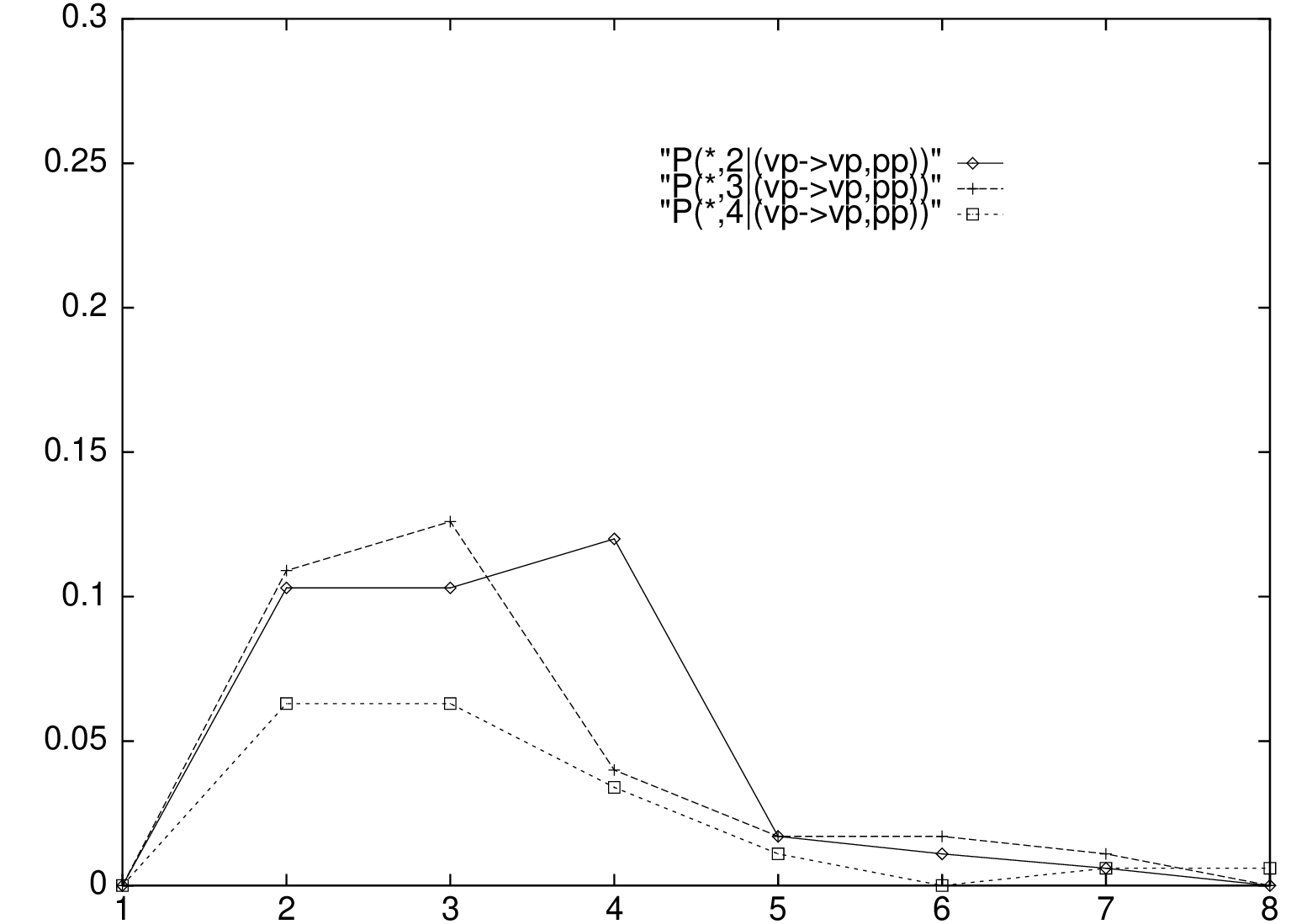}} &
{\epsfxsize6cm\epsfysize4cm\epsfbox{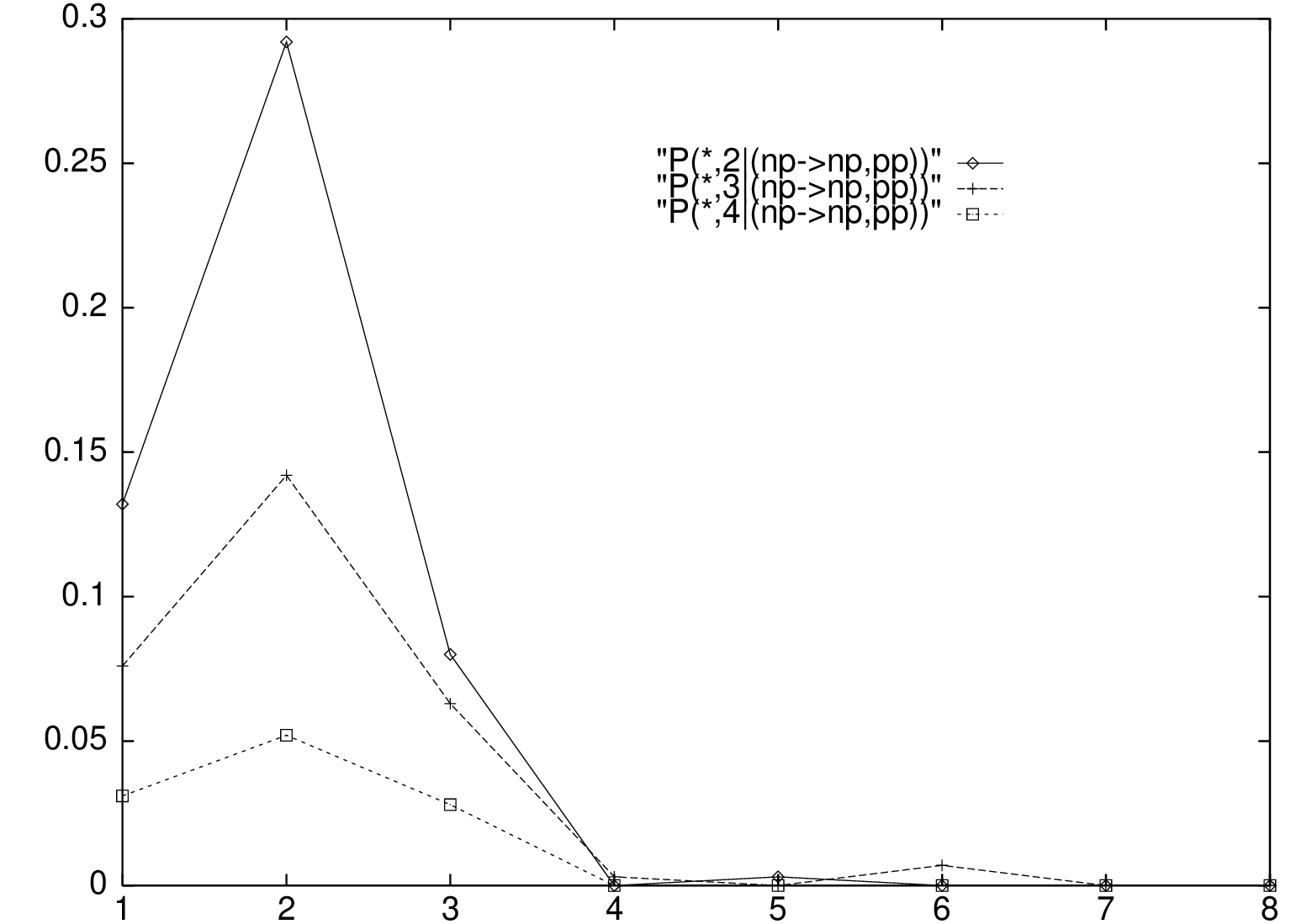}} \\
(a) & (b) \\
\end{tabular}
\caption{Length probability versus length}
\label{fig:exdis}
\end{center}
\end{figure}

We then trained the parameters of probability models. We extracted
$181,250$ case frames from the WSJ (Wall Street Journal) bracketed
corpus of the Penn Tree Bank \cite{Marcus93}. We used these data to
estimate three-word probabilities and two-word probabilities.
Furthermore, we extracted $963$ sentences from the WSJ tagged corpus
of the Penn Tree Bank. We used SAX to analyze the sentences and
selected the {\em correct} syntactic trees by hand. We then employed
the Maximum Likelihood Estimator to estimate length probabilities
using the selected syntactic trees, e.g., if CFG rule $NP \rightarrow
NP,PP$ is applied $x$ times, and among the attachments obtained by
applying this rule, $x_i$ of them have the lengths of $2$ and $3$,
then the length probability $P(2,3|(NP \rightarrow NP,PP))$ is
estimated as $\frac{x_i}{x}$. It is known, in statistics, that the
number of samples required for accurate estimation of a probabilistic
model is roughly proportional to the number of parameters in the
target model, and thus the data used for training length probabilities
were nearly sufficient. Figure~\ref{fig:exdis} plots the estimated
length probabilities versus the lengths, for two CFG rules. The result
indicates that there are more attachments attached to nearby phrases
than are attached to distant ones in the training data. Moreover, the
length probabilities for CFG rule $VP \rightarrow VP,PP$ and those for
CFG rule $NP \rightarrow NP,PP$ show different distribution patterns,
suggesting that syntactic preference is a function of a CFG rule.

We then extracted $249$ sentences from a part of the tagged WSJ corpus
which was not used in training as our test data and analyzed the
sentences. When analizing a sentence, we rank the obtained
interpretations as follows:
\begin{equation}
\begin{array}{llll}
\mbox{if} & P_{lex3}(I_1) > P_{lex3}(I_2) & \mbox{then} & I_1 > I_2 \\
\mbox{else if} & P_{lex3}(I_2) > P_{lex3}(I_1) & \mbox{then} & I_2 >
I_1 \\
\mbox{else if} & P_{lex2}(I_1) > P_{lex2}(I_2) & \mbox{then} & I_1 > I_2 \\
\mbox{else if} & P_{lex2}(I_2) > P_{lex2}(I_1) & \mbox{then} & I_2 > I_1 \\
\mbox{else if} & P_{syn}(I_1) > P_{syn}(I_2) & \mbox{then} & I_1 > I_2 \\
\mbox{else if} & P_{syn}(I_2) > P_{syn}(I_1) & \mbox{then} & I_2 > I_1 \\
\end{array}
\end{equation}
where $I_1$ and $I_2$ denote any two interpretations. $P_{lex3}()$
denotes the lexical likelihood value of an interpretation calculated
as the geometric mean of three-word probabilities, $P_{lex2}()$ the
lexical likelihood value of an interpretation calculated as the
geometric mean of two-word probabilities, and $P_{syn}()$ the
syntactic likelihood value of an interpretation. The average number of
interpretations obtained in the analysis of a sentence was $2.4$.
\begin{table}[htb]
\caption{Disambiguation results}
\label{tab:result}
\begin{center}
\begin{tabular}{|l|c|} \hline
 Method & Accuracy($\%$) \\ \hline 
 Lex3+Lex2+Syn & $89.2$ \\ 
 Lex3+Lex2+PCFG & $86.7$ \\ 
 Lex3$($Lex2$)\times$Syn & $87.1$ \\
 \hline
\end{tabular}
\end{center}
\end{table}
\begin{table}[htb]
\caption{Breakdown of `Lex3+Lex2+Syn'}
\label{tab:breakdown}
\begin{center}
\begin{tabular}{|l|c|c|c|} \hline
& Correct & Incorrect & Total \\ \hline
Lex3 & $112$ & $5$ & $117$ \\
Lex2 & $94$ & $14$ & $108$ \\
Syn & $16$ & $8$ & $24$ \\
Total & $222$ & $27$ & $249$ \\ \hline
\end{tabular}
\end{center}
\end{table}
\begin{figure}[htb]
\begin{center}
{\epsfxsize6cm\epsfysize4cm\epsfbox{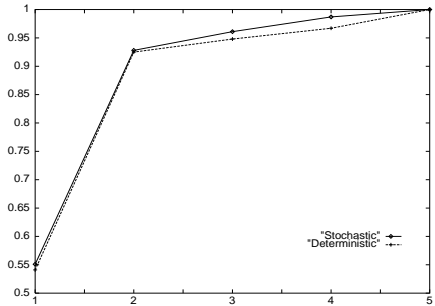}}
\caption{The top 5 accuracies}
\label{fig:acc}
\end{center}
\end{figure}

The number $1$ accuracy obtained was $89.2\%$ (Table~\ref{tab:result}
represents this result as `Lex3+Lex2+Syn'), where the number $n$
accuracy is defined as the fraction of the test sentences whose
preferred interpretation is successfully ranked in the first $n$
candidates. We feel that this result is very encouraging.
Table~\ref{tab:breakdown} shows the breakdown of the result, in which
`Lex3' stands for the proportion determined by using lexical
likelihood $P_{lex3}$, `Lex2' by using lexical likelihood $P_{lex2}$,
and `Syn' by using syntactic likelihood $P_{syn}$. The accuracies of
`Lex3,' `Lex2,' and `Syn' were $95.7\%$, $87.0\%$, and $66.7\%$,
respectively. Furthermore, `Lex3,' `Lex2,' and `Syn' formed $47.0\%$,
$43.4\%$, and $9.6\%$ of the disambiguation results, respectively.

We further examined the types of mistakes made by our method. First,
there were some mistakes by `Syn.' For example, in
\begin{equation}
\mbox{Rain washes the fertilizers off the land,}
\end{equation}
there are two interpretations. The lexical likelihood values
$P_{lex3}$ of the two interpretations were calculated as $0$, and the
lexical likelihood values $P_{lex2}$ of the two interpretations were
calculated as $0$, as well. The interpretations were ranked by the
syntactic likelihood $P_{syn}$, and the interpretation of attaching
the `off' phrase to `fertilizer' was mistakenly preferred. We also
found some mistakes by `Lex2.' For example, in
\begin{equation}
 \mbox{The parents reclaimed the child under the circumstances,}
\end{equation}
there are two interpretations. The lexical likelihood values
$P_{lex3}$ of the two interpretations were calculated as $0$. The
lexical likelihood value $P_{lex2}$ of the interpretation of attaching
`under' phrase to `child' was higher than that of attaching it to
`reclaim,' as there were many expressions like `a child under five'
observed in the training data. And thus the former interpretation was
mistakenly preferred. It is obvious that these kinds of mistakes could
be avoided if more data were available. We conclude that the most
effective way of improving disambiguation results is to increase data
for training lexical preference.

We further checked the disambiguation decisions made by `Syn' when
`Lex3' and `Lex2' fail to work, and found that all of the
prepositional phrases in these sentences were attached to nearby
phrases by `Syn,' indicating that using syntactic likelihood can help
to achieve a functioning of RAP. One may argue that we could obtain
the same number $1$ accuracy if we were to employ a deterministic
approach in implementing RAP. As we pointed out earlier, however, if
we are to obtain the $N$ most preferred interpretations, we need to
use syntactic likelihood. To verify that the syntactic likelihood is
indeed useful, we conducted the following additional experiment. We
ranked the interpretations of each of the $249$ test sentences using
only syntactic likelihood. We also selected the interpretation with
phrases always attached to nearby phrases as the most preferred ones,
and randomly selected interpretations from what remain as the $n$th
most preferred ones. We evaluated the results on the basis of the
number $n$ accuracy. Figure~\ref{fig:acc} shows the top $5$ accuracies
of the stochastic approach and the deterministic approach. The results
indicate that the former outperforms the latter. (The number $2$
accuracy for both methods increases drastically, as many test
sentences have only two interpretations.) The improvement is not
significant, however. We expect the effect of the use of the syntactic
likelihood to become more significant when longer sentences are used
in future analyses.

In place of a length probability model, we used PCFG for calculating
syntactic preference. We employed the Maximum Likelihood Estimator to
estimate the parameters of PCFG (we did not use the so-called
`inside-outside algorithm' \cite{Jelinek90,Lari90}), making use of the
same training data as those used for the length probability model.
Table~\ref{tab:result} represents this result as `Lex3+Lex2+PCFG.'
Our experimental results indicate that our method of using a length
probability model outperforms that of using PCFG.

Instead of the back-off method, we used the product of lexical
likelihood values and syntactic likelihood values to rank
interpretations. When using lexical likelihood, we use a lexical
likelihood value calculated from three-word probabilities, provided
that it is not $0$; otherwise we use a lexical likelihood value
calculated from two-word probabilities. Table~\ref{tab:result}
represents this result as `Lex3(Lex2)$\times$Syn.' When the preference
values of all of the interpretations obtained are calculated as $0$,
we rank the interpretations at random. Our results indicate that it is
preferable to employ the back-off method.
\section{Concluding Remarks}
We have proposed a probabilistic method of disambiguation based on
psycholinguistic principles. Our main proposals are: (a) to unify the
psycholinguistic approach and the probabilistic approach,
specifically, to implement psycholinguistic principles on the basis of
probabilistic methodology, (b) to use the notion of `length' in
defining a probabilistic model for the implementation of RAP and ALPP,
and (c) to employ the back-off method to combine the use of lexical
likelihood with that of syntactic likelihood. Our experimental results
indicate that our method is quite effective.

\section*{Acknowledgement}

I thank greatly Mr.~K.~Nakamura, Mr.~T.~Fujita, and Dr.~K.~Kobayashi
of NEC for their constant encouragement. I also thank greatly
Dr.~N.~Abe of NEC, and Dr.~Y.~Den of ATR for their valuable comments
and suggestions.

\end{document}